\documentclass[runningheads]{llncs}

\usepackage{accv}

\usepackage{accvabbrv}

\usepackage{graphicx}
\usepackage{booktabs}

\usepackage[accsupp]{axessibility}  

\usepackage{hyperref}

\usepackage{orcidlink}
\usepackage{caption}

\usepackage{graphicx}

\usepackage{comment}
\usepackage{multirow}
\usepackage{pgfplots}
\usetikzlibrary{spy,calc}
\usepackage{multirow}

\usepackage[capitalize]{cleveref}
\crefname{section}{Sec.}{Secs.}
\Crefname{section}{Section}{Sections}
\Crefname{table}{Table}{Tables}
\crefname{table}{Tab.}{Tabs.}

\begin{document}

\title{PixMamba: Leveraging State Space Models in a Dual-Level Architecture for Underwater Image Enhancement} 

\titlerunning{PixMamba}
%
\author{Wei-Tung Lin\inst{1}
\quad
Yong-Xiang Lin\inst{1}
\quad
Jyun-Wei Chen\inst{1}
\quad
Kai-Lung Hua\inst{1,2}
}
\authorrunning{W. Lin \etal}
%
\institute{Dept. of Computer Science and Information Engineering,\\National Taiwan University of Science and Technology\\
\email{\{m11115116,d10915006,m11215034,hua\}@mail.ntust.edu.tw}
\and
Microsoft Taiwan\\
\email{kai.hua@microsoft.com}
}

\maketitle

\begin{abstract}

Underwater Image Enhancement (UIE) is critical for marine research and exploration but hindered by complex color distortions and severe blurring. Recent deep learning-based methods have achieved remarkable results, yet these methods struggle with high computational costs and insufficient global modeling, resulting in locally under- or over-adjusted regions. We present PixMamba, a novel architecture, designed to overcome these challenges by leveraging State Space Models (SSMs) for efficient global dependency modeling. Unlike convolutional neural networks (CNNs) with limited receptive fields and transformer networks with high computational costs, PixMamba efficiently captures global contextual information while maintaining computational efficiency. Our dual-level strategy features the \textbf{patch-level} Efficient Mamba Net (EMNet) for reconstructing enhanced image feature and the \textbf{pixel-level} PixMamba Net (PixNet) to ensure fine-grained feature capturing and global consistency of enhanced image that were previously difficult to obtain.  PixMamba achieves state-of-the-art performance across various underwater image datasets and delivers visually superior results. Code is available at: \url{https://github.com/weitunglin/pixmamba}.



\end{abstract}

\section{Introduction}

Underwater environments pose unique challenges for image acquisition due to factors such as severe blurring, color distortion\cite{ancuti2018color}, low contrast, and complex light scattering\cite{peng2017underwater,li2016single} caused by wavelength-dependent absorption. These issues impede the quality and clarity of underwater images, making effective enhancement methods critical for various applications in marine archaeology, ecological and biological research. Therefore, underwater image enhancement (UIE) is a crucial step in improving the underwater images quality. This enhancement facilitates improved understanding of the underwater world and enables the successful execution of high-level oceanography tasks\cite{cao2023dynamic,fayaz2024intelligent}.

Traditional image enhancement methods\cite{ancuti2018color,berman2021underwater,zhuang2022underwater} have relied on statistical properties and physical assumptions about the image and environment. These methods often attempt to correct color distortions and improve contrast using hand-crafted priors. However, they typically struggle with dynamic scenes and often fall short in restoring texture information and handling extensive blurring. Recent advancements in deep learning have introduced new approaches to underwater image enhancement. Methods utilizing convolutional neural networks (CNNs) are widely used for UIE due to their capability to learn visual representations end-to-end\cite{li2020uieb, li2021ucolor,fu2022puienet,cong2023PUGAN,huang2023semiuir}, which is more efficient and effective compared to traditional UIE methods. However, CNN-based methods have limitations: a small receptive field hinders modeling long-range pixel dependencies, and fixed convolutional kernels cannot adapt to the images across various underwater scenarios. The Transformer-based model, initially proposed for natural language processing\cite{vaswani2017attention} and further applied to vision tasks\cite{liu2021swin}, could overcome the limitations of CNNs and archives remarkable performance results. However, quadratic complexity with respect to sequence length of Transformer poses a serious problem for its application in real-world underwater image enhancement (UIE) scenarios that may requires processing high-resolution images in real-time efficiency.

Recently, State Space Models (SSM) and their improved variants, Mamba\cite{gu2023mamba} and Mamba-2\cite{tri2024mamba2}, have emerged as efficient and effective backbones for long-sequence modeling. This evolution hints at a potential solution for balancing global receptive fields and computational efficiency for computer vision tasks. The discretized state-space equations in Mamba can be formalized into a recursive form, enabling the modeling of very long-range dependencies through specially designed structured reparameterization. This capability allows Mamba-based restoration networks learns and interprets the images context better, thereby enhancing reconstruction quality\cite{guo2024mambair}. Additionally, Mamba's parallel scan algorithm facilitates the parallel processing of each token, making efficient use of modern hardware like GPUs. These promising properties motivate us to explore the potential of Mamba-based architecture for achieving both efficient and effective in image restoration tasks.

Given the challenges in underwater image enhancement, we present PixMamba, a novel approach that utilizes the linear complexity and long-range modeling capabilities of State Space Models (SSMs). PixMamba is tailored for efficient and effective underwater image enhancement, consisting of two key components operating at different levels: the Efficient Mamba Net (EMNet) and the PixMamba Net (PixNet). EMNet combines the Efficient Mamba Block (EMB) for efficient patch-level feature extraction and dependency modeling with the Mamba Upsampling Block (MUB) for detail-preserving upsampling. However, relying solely on patch-level processing can lead to inconsistencies and fail to capture long-range dependencies that govern overall clarity, color balance, and global consistency. 

To address this limitation, PixMamba introduces PixNet, a parallel pixel-level network that processes the entire image at the pixel level, capturing detailed features at global view. This hybrid dual-level architecture allows PixMamba to leverage the strengths of both patch-level and pixel-level processing. EMNet extracts rich localized features within small patches for reconstructing high-quality images, while PixNet efficiently models interactions and dependencies at the individual pixel level across the entire image. By capturing these global relationships, PixNet recovers overall aesthetic qualities often degraded in underwater conditions, such as contrast, saturation, and haze removal. Combining these complementary levels enables PixMamba to simultaneously enhance microscopic details and macroscopic image qualities. EMNet preserves intricate textures and structures, while PixNet ensures global consistency, clarity, and natural-looking results. PixMamba effectively adapts the promising performance of SSMs into the image restoration task, achieving state-of-the-art results on various underwater datasets. To the best of our knowledge, it is the first approach to process the entire image at the pixel level, a capability made feasible by the linear complexity of SSMs, as opposed to the quadratic complexity of traditional transformer-based methods. Existing methods relying solely on skip connections at the final layer have proven insufficient for high-quality image restoration, as demonstrated by our empirical results. PixMamba's hybrid dual-level architecture addresses these limitations, leveraging the advantages of both patch-level and pixel-level processing for superior underwater image enhancement.

Compared to existing SSM-based UIE method\cite{guan2024watermamba} only utilizes patch-level processing, which may potentially loss detailed features and lead to global inconsistency of the enhanced image. Our proposed dual-level processing enables PixMamba to capture fine-grained feature and ensure overall consistency and clarity.

In this paper, we present the following key contributions:
\begin{itemize}

\item \textbf{PixMamba:} A novel dual-level architecture, PixMamba introduces a novel approach to highly efficient and detailed image restoration. By seamlessly integrating local patch-level processing through Efficient Mamba Net (EMNet) and global pixel-level processing via the innovative PixMamba Net (PixNet), PixMamba delivers refined and enhanced high-quality underwater images.

\item \textbf{EMNet:} As a core component of PixMamba, EMNet adeptly combines the Efficient Mamba Block (EMB) and Mamba Upsampling Block (MUB). EMB excels in capturing essential image features with better memory efficiency, while MUB specializes in preserving intricate details during the upsampling process. This combination significantly improves the quality of restored images and enhances overall processing efficiency.

\item \textbf{State-of-the-art Performance}: The synergy between PixNet's pixel-level feature extraction and EMNet's robust patch-level processing enables PixMamba to achieve a more refined and enhanced underwater image restoration process, reaches remarkable results across various UIE datasets.

\end{itemize}

\section{Related Works}

Traditional physical-based and prior-based methods for underwater image enhancement are increasingly being replaced by deep learning-based approaches due to their superior ability to learn feature representations from underwater images through the deep learning process. Compared to traditional hand-crafted algorithms, deep learning-based methods have gained more interests. Deep learning-based image enhancement approaches have three main categories: CNN-based, Transformer-based, and the most recent SSM-based. Each category will be discussed in the following section respetively.

\subsection{CNN-based Image Enhancement}

Li \etal\cite{li2021ucolor} introduced a network that employs an embedding strategy spanning multiple color spaces, guided by transmission properties. Their approach utilizes an encoder that combines different color space representations and a decoder that enhances degraded regions based on transmission guidance. Fu \etal\cite{fu2022puienet} models UIE into a distribution estimation problem. It first used a probabilistic network based on a conditional VAE and adaptive instance normalization that learns to approximate the posterior over meaningful appearance. Cong \etal\cite{cong2023PUGAN} proposed a physical model-guided Generative Adversarial Network (GAN) for UIE. The network incorporates a Parameters Estimation subnetwork for learning physical model parameters and a Two-Stream Interaction Enhancement subnetwork with a Degradation Quantization module for key region enhancement, along with Dual-Discriminators for style-content adversarial constraints. Huang \etal\cite{huang2023semiuir} developed a Semi-supervised Underwater Image Restoration framework (Semi-UIR) based on the mean-teacher model. To address limitations with the naive approach, they introduced a reliable bank for pseudo ground truth and incorporated contrastive regularization to combat confirmation bias. However, CNNs\cite{lin2019spatially,shahid2021spatio,shahid2021deep} suffer from the inherent limitation of the local receptive field mechanism and insufficient to learn global representations.

\subsection{Transformer-based Image Enhancement}

Ren \etal\cite{ren2022urstc} proposed a novel approach using the U-Net based Reinforced Swin-Convs Transformer. By embedding Swin Transformer into U-Net, they enhanced the model's ability to capture global dependencies while reintroducing convolutions to capture local attention. Zamir \etal\cite{zamir2021restormer} introduced an efficient Transformer model, designed to handle high-resolution image restoration tasks. It makes strategic modifications to the multi-head attention and feed-forward network modules, enabling it to capture long-range pixel interactions while being computationally manageable. Gu \etal\cite{gu2022convformer} presented a hierarchical CNN and Transformer hybrid architecture. This architecture includes a residual-shaped hybrid stem combining convolutions with an Enhanced Deformable Transformer (DeTrans), capable of learning both local and global representations and exploiting multi-scale features effectively. Nervertheless, self-attention mechanism in Transforms\cite{chen2024transformer} scales massively for high-resolution images, which is impractical for real-world applications.

\subsection{SSM-based Image Enhancement}

Guan \etal\cite{guan2024watermamba} introduce a state space model (SSM) for UIE that aims to combine linear computational complexity with effective degradation handling. To address spatial and channel dependencies, their model includes spatial-channel omnidirectional selective scan blocks and multi-scale feedforward networks to promote coordinated information flow and adjust image details. By utilizing SSM, superior performance has been shown in addressing the limitations of CNN's poor generalizability and the Transformer's computational inefficiency.
\section{Methods}
\subsection{Preliminaries}
Structured State Space Models (S4) is a recent sequence model which related to RNNs, CNNs, and classical state space model. This model, inspired by continuous systems, essentially maps a one-dimensional sequence $x(t) \in \mathbb{R}$ to an output sequence $y(t) \in \mathbb{R}$ via hidden state $h(t) \in \mathbb{R}^N$. Continuous systems can be formulated in linear ordinary differential equation (ODE) as follows:
\begin{align}
h'(t) &= A h(t) + B x(t) \label{eq1} \\
y(t) &= Ch(t) + Dx(t) \label{eq2}
\end{align}
where $h(t) \in \mathbb{R}^N$ is the hidden state and $A \in \mathbb{R}^{N \times N}, B \in \mathbb{R}^N$ and $C \in \mathbb{R}^N$ are the parameters when the state size is equal to $N$. $D \in \mathbb{R}$ represents the skip connection.

Next, it is necessary to discretize \eqref{eq1} and \eqref{eq2}. The discretized form of \eqref{eq1} can be obtained using the zero-order hold (ZOH) rule, which requires $A$ and $B$ to be converted into discrete forms using the time scaling parameter $\Delta$. Consequently, the discretization can be defined as follows:
\begin{align}
h'(t) &= \overline{A} h_{t-1} + \overline{B} x_t \\
y(t) &= C h_t + D x_t \\ 
\overline{A} &= e^{\Delta A} \\
\overline{B} &= (\Delta A)^{-1}(e^{\Delta A} - I)
\end{align}
where $\Delta \in \mathbb{R}^D$ is the time scale parameter and $B, C \in \mathbb{R}^{D\times N}$.

\subsection{Overall Architecture}
\begin{figure}[!t]
    \centering
    \includegraphics[width=1\linewidth]{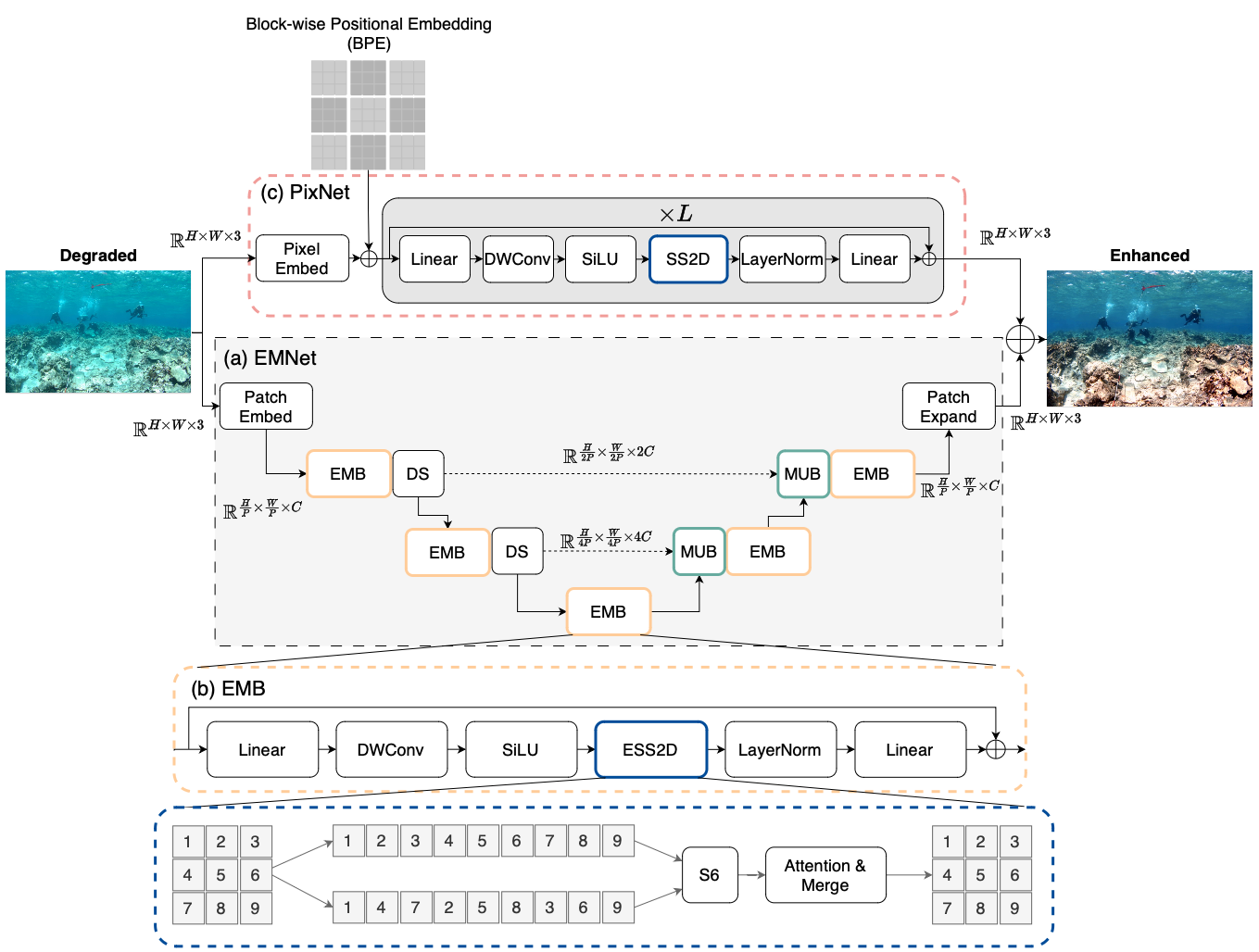}
    \caption{Overall architecture of PixMamba. EMNet: Efficient Mamba Net; EMB: Efficient Mamba Block; PixNet: PixMamba Net; MUB: Mamba Upsampling Block; DS: Downsampling Block; DWConv: Depth-wise Convolution Block; S6: Mamba SSM\cite{gu2023mamba}.}
    \label{fig:arch}
\end{figure}

The overall architecture of our concept, PixMamba, relies on a Efficient Mamba Net (EMNet) architecture and integrates a PixMamba Net (PixNet) in parallel as shown in \cref{fig:arch}. Given a degraded underwater image, $I \in \mathbb{R}^{H \times W \times 3}$, EMNet begins with encoding via PatchEmbed, generating $I^0_E \in \mathbb{R}^{\frac{H}{P} \times \frac{W}{P} \times D}$. The patched image $I^0_E$ undergoes further encoding across three stages via the Efficient Mamba Block (EMB), with a downsampling Layer applied after each stage, resulting in downsampled image feature sizes of $I^1_E \in \mathbb{R}^{\frac{H}{\textit{2P}}\times\frac{W}{\textit{2P}}}$ and $I^2_E \in \mathbb{R}^{\frac{H}{\text{4P}}\times\frac{W}{\text{4P}}}$, respectively. Following this, the Mamba Upsampling Block (MUB) and EMB are applied in three stages subsequently to decode the features and obtain $I^2_D$, $I^1_D$, and $I^0_D$ with sizes of $\frac{H}{\text{4P}}\times\frac{W}{\text{4P}}$, $\frac{H}{\text{2P}}\times\frac{W}{\text{2P}}$, and $\frac{H}{P}\times\frac{W}{P}$, respectively,. From $I^0_D \in \mathbb{R}^{\frac{H}{P} \times \frac{W}{P} \times D}$, we then expand and project to form $I_{\text{FD}} \in \mathbb{R}^{H \times W \times 3}$. 

To incorporate the global detailed pixel-level information into our framework, PixNet is introduced. This network contains $L$ stages via Mamba Block, sequentially enhancing the deep features of the underwater image $I^l_P \in \mathbb{R}^{\text{HW} \times D}$ layer by layer, where $l \in \{1, 2, \ldots, L\}$. Specifically, PixNet embeds $I$ pixel-wise into a feature $I^0_P \in \mathbb{R}^{\text{HW} \times D}$ and enriches the local information of image with bi-linear sampled Block-wise Positional Embedding (BPE), $\text{BPE} \in \mathbb{R}^{\frac{\text{HW}}{B^2} \times D}$. At the final stage, the deep features $I^L_P$ are projected to $I_{\text{FP}} \in \mathbb{R}^{H \times W \times 3}$. 
After obtaining the pixel-level detailed information ($I_{\text{FP}}$) and the spatially rich patch-level information ($I_{\text{FD}}$), we combine them to form our refined enhanced underwater image, $I_F = I_{\text{FD}} + I_{\text{FP}},  I_F \in \mathbb{R}^{H \times W \times 3}$, where $B$ is block size of $\text{BPE}$, $P$ is patch size, $H, W$ are image height and width, $D$ is hidden dimension, $I^l_E, I^l_D$ are encoded and decoded feature of EMNet at $l$ layer, and $I^l_P$ is decoded feature of PixNet at $l$ layer.

\subsection{EMNet} \label{sec:emnet}
Our proposed EMNet, depicted in \cref{fig:arch}(a), integrates the remarkable ability of the SSM to capture both global and local feature dependencies into the proven successful U-Net architecture for image restoration\cite{ronneberger2015unet}. While naively integrating SSM into a U-Net architecture significantly increases computational complexity due to the doubling of the hidden dimension size at each stage. Therefore, EMNet reduces the number of stages in U-Net by one to efficiently save memory and computational cost. Furthermore, we have doubled the patch size of the initial stage to maintain the same receptive field as the original U-Net design, promoting better restoration performance. Building upon this efficient U-Net architecture, EMNet further incorporates two core components: the Efficient Mamba Block (EMB) and Mamba Upsampling Block (MUB).

\subsubsection{Efficient Mamba Block} \label{EMB}
As illustrated in \cref{fig:arch}(b), the Efficient Mamba Block processes image feature patches using Efficient SS2D (ESS2D). This block is a more computationally efficient variant of the 2D selective scan operation introduced by VMamba\cite{liu2024vmamba}. While the original SS2D utilizes four-direction scans to model dependencies between patches, ESS2D simplifies this process to significantly reduce computational cost with minimal performance loss. This simplification maintains the effectiveness of modeling feature dependencies while improving efficiency.

Following the independent processing of features within each scanning branch, a spatial and channel attention module\cite{huang2024localmamba} is employed to combine the extracted information and remove channel redundancy. This attention module\cite{huang2024localmamba} refines the feature representation by adjusting in both channel-wise and spatial-wise. It comprises two branches: a channel attention branch that captures broad feature representations and assigns weights across channels, and a spatial attention branch that assesses the significance of individual tokens within the features, allowing for detailed and importance-weighted feature extraction. Finally, after this attention-based filtering, features from the distinct scanning branches are merged to create the final deep feature representation.

\subsubsection{Mamba Upsampling Block} \label{MUB}
The U-Net architecture utilizes a symmetric encoder-decoder style, where feature extraction is performed through downsampling, and upsampling is used to restore the image's original features at different scales. However, the traditional U-Net architecture's upsampling process can suffer from drawbacks such as the loss of fine details or noise introduction, leading to sub-optimal performance in some scenarios.

To address these challenges, we propose an Mamba Upsampling Block that incorporates a SSM mechanism before the upsampling process. SSM has the ability to effectively capture the dependencies among channels and tokens. By integrating SSM prior to the upsampling stage, we can intelligently adjust the features that required for upsampled, selectively maintaining and prioritizing the most relevant information within the feature maps. This innovative approach aims to significantly improve detail preservation during upsampling, thereby enhancing the overall performance and the quality of the restored image. This process is defined as follows:
\begin{align}
I^{s-1}_{D} &= \textbf{Norm}(\textbf{TransposeConv2D}(\textbf{EMB}(I^s_D W)))
\end{align}
where $W$ is learnable projection matrix, $I^s_D$ is decoded feature of EMNet at $s$-th stage.

\subsection{PixMamba Net}
We propose a novel concept named PixMamba Net (PixNet), illustrated in \cref{fig:arch}(c),that operates at the pixel-level to capture finer details within an image. Compared to EMNet(\cref{sec:emnet}), which performs operations at a patch-level (such as 2x2 or 4x4 pixel patches), PixNet is specifically designed to perform processing at a pixel-level that can potentially lead to enhanced images with sharper details and improved noise reduction. At the core of the PixNet module is  Mamba Block, where SSM is employed at the pixel-level. With SSM, it exploits the full potential of each individual pixel rather than merely relying on the patch-level information, enabling the model to acquire more fine-grained features and global consistency from the original image. By leveraging PixNet's pixel-level processing, we can attain improved information extraction, global consistency and overall image clarity. It significantly increasing the performance and the quality of the image enhancement. This innovation propels the limits of what is achievable with U-Net architecture.

To enable PixNet to have local spatial information alongside with its global pixel-level information, Block-wise learning Positional Embedding (BPE) is introduced to address this challenge. Having spatial information is essential for SSM modeling due to the natural property of sequential modeling, which flattens all tokens into 1-dimension sequence. Thereby, BPE has block-wise design, which split the positional embedding into $\frac{HW}{B^2}$, and further bi-linear sampled into input sequence size. Finally, BPE is added into the pixel-level sequence feature before the PixNet processing to provide the spatial information for more effective pixel-level sequence processing. The entire process is defined as follows:
\begin{align}
\text{PE} &= \textbf{Upsample}(\text{BPE}) \\
I^0_P &= [I^0 W ; I^1 W ; \ldots ; I^{HW} W ] + \text{PE} \\
I^l_P &= \textbf{MambaBlock}(I^{l-1}_P) + I^{l-1}_P \\
I_{\text{FP}} &= \textbf{Project}(I^L_P) 
\end{align}
where $W \in \mathbb{R}^{HW \times D}$ is the
learnable projection matrix, and $I^i$ is $i$-th pixel of input image $I$.
\section{Experiments}
\label{sec:exp}

\subsection{Implementation Details}

The proposed PixMamba was built using the PyTorch 2.1.0 and MMagic\cite{mmagic2023} toolkits. We used an NVIDIA RTX 3090 GPU for all training and testing experiments. The model was trained end-to-end using the Charbonnier Loss\cite{lai2017LapSRN} and the AdamW\cite{loshchilov2018decoupled} optimizer. The learning rate was set to $4e^{-4}$, with $\beta_1=0.9$ and $\beta_2=0.99$. All images were resized to $256 \times 256$ pixels. Training batch size was set to 16, and PixMamba network was trained for 800 epochs. Learning rate was adjusted using a 20-epoch warm-up, followed by a cosine annealing scheduler\cite{loshchilov2017sgdr}.

\begin{table}[h]
	\centering
    \caption{{Quantitative comparisons across \textbf{C60} and \textbf{UCCS} datasets, model parameters, and FLOPs. Best highlighted in \textbf{bold} and second in \underline{underline}.}}
	\fontsize{7.5}{12}\selectfont
	\setlength{\tabcolsep}{0.8mm}{
	\begin{tabular}{ccccccccc}
		\hline
		 \multirow{2}{*}{ \textbf{Method}} & \multirow{2}{*}{\textbf{Venue}}&  \multicolumn{2}{c}{\textbf{C60}}& \multicolumn{2}{c}{\textbf{UCCS}}&\multirow{2}{*}{ \textbf{Params} $\downarrow$ }&\multirow{2}{*}{ \textbf{FLOPs} $\downarrow$ }
\\&& UIQM $\uparrow$ & UCIQE $\uparrow$ & UIQM $\uparrow$ & UCIQE $\uparrow$ \\
		\hline

		Ucolor\cite{li2021ucolor} & TIP 21 & 2.482&  0.553 &3.019 &0.550 & 157.4M &34.68G \\
		PUIE-Net\cite{fu2022puienet} & ECCV 22 &2.521 &0.558 &3.003 &0.536  & \textbf{1.41M} &30.09G \\
        URSCT\cite{ren2022urstc} & TGRS 22 &2.642 &0.543 &2.947 &0.544&11.41M &18.11G\\
		Restormer\cite{zamir2021restormer} & CVPR 22 &2.688 &0.572 &2.981 &0.542  &26.10M &140.99G \\

		PUGAN\cite{cong2023PUGAN} & TIP 23 &2.652 &0.566 &2.977 &0.536 &95.66M &72.05G  \\

		MFEF\cite{zhou2023mfef} & EAAI 23 &2.652 &0.566 &2.977 &\underline{0.556} &61.86M &26.52G \\
		Semi-UIR\cite{huang2023semiuir} & CVPR 23 &2.667&0.574 & \textbf{3.079} &0.554 & \underline{1.65M} &36.44G \\

		Convformer\cite{gu2022convformer} & TETCI 24 &2.684 &0.572 &2.946 &0.555 &25.9M &36.9G \\
		X-CAUNET\cite{pramanick2024xcaunet} & ICASSP 24 &2.683 &0.564 &2.922 &0.541 &31.78M &261.48G \\

		WaterMamba\cite{guan2024watermamba} & arXiv 24 &\underline{2.853}  &\underline{0.582} & \underline{3.057} &0.55&3.69M &\textbf{7.53G} \\
		\hline

        PixMamba (Ours) & - & \textbf{2.868} & \textbf{0.586} & 3.053 & \textbf{0.561} & 8.68M & \underline{7.60G} \\ 

        \hline
		\end{tabular}
	}

    \vspace{6pt}

    \label{table_c60_uccs}    
    
\end{table}
\begin{figure}[!tb]
    \centering
    \begin{subfigure}[b]{0.24\linewidth}
        \centering
        \begin{tikzpicture}[spy using outlines={circle,red,magnification=2,size=.3\textwidth, connect spies}]
        \node {\includegraphics[width=1\textwidth,
        keepaspectratio]{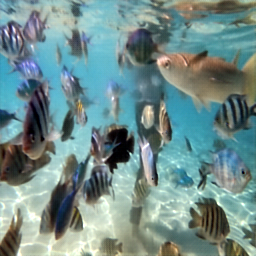}};
        \spy on (1.2,0.8) in node [left] at (1.1,-1);
        \spy on (-0.5,0.2) in node [left] at (-0.5,-1);
        \end{tikzpicture}
        \caption{WaterMamba\cite{guan2024watermamba}}
    \end{subfigure}
    \begin{subfigure}[b]{0.24\linewidth}
        \centering
        \begin{tikzpicture}[spy using outlines={circle,red,magnification=2,size=.3\textwidth, connect spies}]
        \node {\includegraphics[width=1\textwidth,
        keepaspectratio]{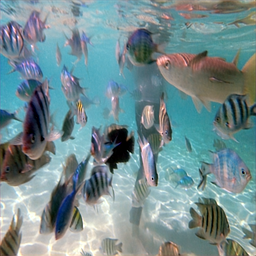}};
        \spy on (1.2,0.8) in node [left] at (1.1,-1);
        \spy on (-0.5,0.2) in node [left] at (-0.5,-1);
        \end{tikzpicture}
        \caption{Semi-UIR\cite{huang2023semiuir}}
    \end{subfigure}
    \begin{subfigure}[b]{0.24\linewidth}
        \centering
        \begin{tikzpicture}[spy using outlines={circle,red,magnification=2,size=.3\textwidth, connect spies}]
        \node {\includegraphics[width=1\textwidth,
        keepaspectratio]{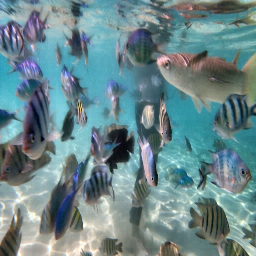}};
        \spy on (1.2,0.8) in node [left] at (1.1,-1);
        \spy on (-0.5,0.2) in node [left] at (-0.5,-1);
        \end{tikzpicture}
        \caption{Ours}
    \end{subfigure}
    \begin{subfigure}[b]{0.24\linewidth}
        \centering
        \begin{tikzpicture}[spy using outlines={circle,red,magnification=2,size=.3\textwidth, connect spies}]
        \node {\includegraphics[width=1\textwidth,
        keepaspectratio]{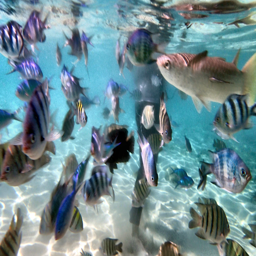}};
        \spy on (1.2,0.8) in node [left] at (1.1,-1);
        \spy on (-0.5,0.2) in node [left] at (-0.5,-1);
        \end{tikzpicture}
        \caption{reference}
    \end{subfigure}
    \caption{Enhanced image detail visualization. Our method improves the detail features of the degraded image compared to WaterMamba\cite{guan2024watermamba} and Semi-UIR\cite{huang2023semiuir} As highlighted in the red circle, our approach shows superior result on the detail features over WaterMamba\cite{guan2024watermamba} and Semi-UIR\cite{huang2023semiuir}, demonstrating the advantage of our proposed MUB and PixNet techniques.}
    \label{fig:detail_comp}
\end{figure}

\begin{table}[h]
    \centering
       \caption{Quantitative comparisons on \textbf{T90} dataset, model parameters, and FLOPs. Best highlighted in \textbf{bold} and
second in \underline{underline}.}
	\fontsize{8}{12}\selectfont
	\setlength{\tabcolsep}{0.8mm}{
	\begin{tabular}{ccccccc}
		\hline
    		 \multirow{2}{*}{\textbf{Method}} & \multirow{2}{*}{\textbf{Venue}} & \multicolumn{5}{c}{\textbf{T90}}
        \\&&PSNR $\uparrow$ &SSIM $\uparrow$&MSE $\downarrow$&UIQM $\uparrow$&UCIQE $\uparrow$ \\
		\hline
		Ucolor\cite{li2021ucolor} & TIP 21 &21.093  &0.872 & 0.096 & \textbf{3.049} & 0.555  \\
	   Shallow-uwnet\cite{naik2021shallowuwnet} & AAAI 21 & 18.278 & 0.855 & 0.131 & 2.942 & 0.544  \\
        UIEC\^{}2-Net\cite{wang2021uiec} & SPIC 21 & 22.958 & 0.907 & 0.078 & 2.999 & 0.599  \\
 
		PUIE-Net\cite{fu2022puienet} & ECCV 22 &21.382 &0.882 &0.093&  3.021 &0.566 \\

        NU$^2$Net\cite{guo2023uranker} & AAAI 23 & 23.061 & \underline{0.923} & 0.086 & 2.936 & 0.587  \\

        FiveA+\cite{jiang2023five} & BMVC 23 & 23.061 & 0.911 & 0.076 & 2.828 & \underline{0.616} \\

		WaterMamba\cite{guan2024watermamba} & arXiv 24 & \textbf{24.715} & \textbf{0.931} & - & - & - \\
  
		\hline

        PixMamba (Ours) & - & \underline{23.587} & 0.921 & \textbf{0.061} & \underline{3.048} & \textbf{0.617} \\

        \hline
		\end{tabular}
	}

    \label{table_t90}    
\end{table}








\subsection{Datasets}

The experiments used two publicly available underwater image datasets: UIEB\cite{li2020uieb} and UCCS\cite{liu2020uccs}. UIEB dataset has total of 950 images, and was split into a train set of 800 samples (U800), a validation set of 90 samples (T90), and a challenge set of 60 samples (C60). Each sample in U800 and T90 includes a raw degraded underwater image and its corresponding human-curated reference image, while C60 has only degraded image\cite{li2020uieb}. The UCCS dataset includes three different underwater color settings: bluish, greenish and blue-green tones, each setting contains 100 images, totaling 300 images\cite{liu2020uccs}.

\subsection{Evaluataion Metrics}

We evaluated our PixMamba method using five criteria. First, there's the Mean Squared Error (MSE) calculates average squared per-pixel error. Then, Peak Signal-to-Noise Ratio (PSNR)\cite{korhonen2012psnr} gauges the ratio between the image's signal to its noise, offering a measure of the overall image quality. The Structural Similarity Index (SSIM)\cite{wang2004ssim} measures how similar the image structure is, which aligns closely with human vision. Underwater Image Quality Measure (UIQM)\cite{panetta2016uiqm} comprises of three underwater image attributed measures: image colorfulness, sharpness, and contrast. Lastly, the Underwater Color Image Quality Evaluation (UCIQE) metric assesses the overall image smoothness, clarity, and contrast. It is worth noting that both UIQM and UCIQE are no-reference evaluation methods, designed to assess the quality of images without the need for a reference.

\begin{figure}[!tb]    
	\centering
	\begin{subfigure}[b]{0.09\linewidth}
			\centerline{\includegraphics[width=1.2cm,height=1.2cm]{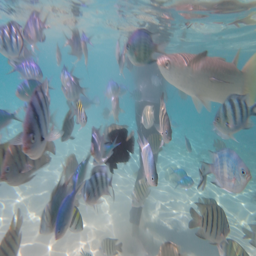}}
			\centerline{\includegraphics[width=1.2cm,height=1.2cm]{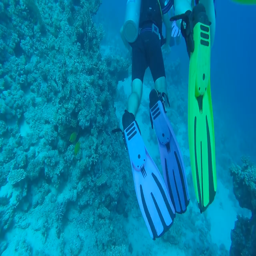}} 

			\centerline{(a)}
	\end{subfigure}\hfill
	\begin{subfigure}[b]{0.09\linewidth}
			\centerline{\includegraphics[width=1.2cm,height=1.2cm]{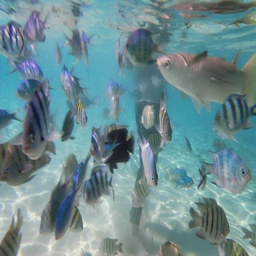}}
			\centerline{\includegraphics[width=1.2cm,height=1.2cm]{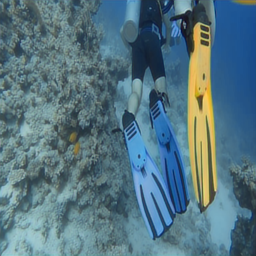}} 

			\centerline{(b)}
	\end{subfigure}\hfill
	\begin{subfigure}[b]{0.09\linewidth}
			\centerline{\includegraphics[width=1.2cm,height=1.2cm]{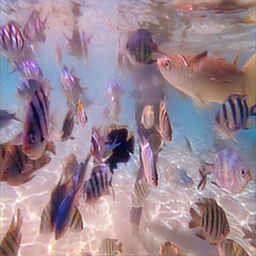}}
			\centerline{\includegraphics[width=1.2cm,height=1.2cm]{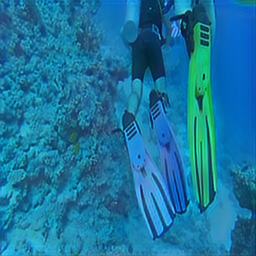}} 
			
			\centerline{(c)}
	\end{subfigure}\hfill
	\begin{subfigure}[b]{0.09\linewidth}
			\centerline{\includegraphics[width=1.2cm,height=1.2cm]{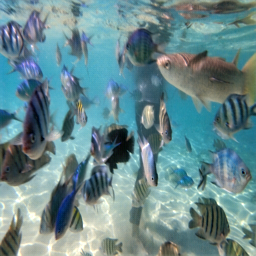}}
			\centerline{\includegraphics[width=1.2cm,height=1.2cm]{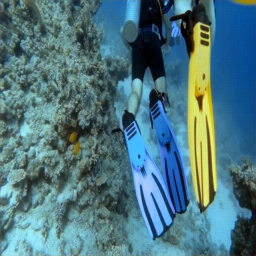}}

			\centerline{(d)}
	\end{subfigure}\hfill
	\begin{subfigure}[b]{0.09\linewidth}
			\centerline{\includegraphics[width=1.2cm,height=1.2cm]{uie/semi/r90/1}}
			\centerline{\includegraphics[width=1.2cm,height=1.2cm]{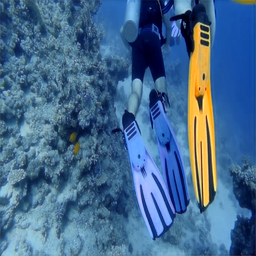}} 
			
			\centerline{(e)}
	\end{subfigure}\hfill
	\begin{subfigure}[b]{0.09\linewidth}
			\centerline{\includegraphics[width=1.2cm,height=1.2cm]{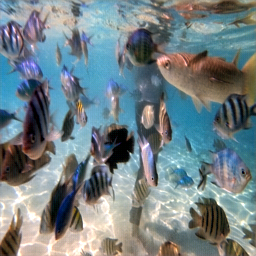}}
			\centerline{\includegraphics[width=1.2cm,height=1.2cm]{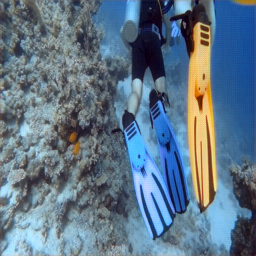}} 
			
			\centerline{(f)}
	\end{subfigure}\hfill
	\begin{subfigure}[b]{0.09\linewidth}
			\centerline{\includegraphics[width=1.2cm,height=1.2cm]{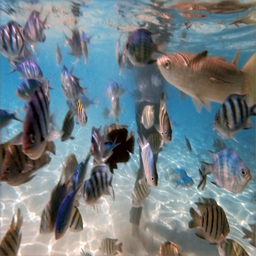}}
			\centerline{\includegraphics[width=1.2cm,height=1.2cm]{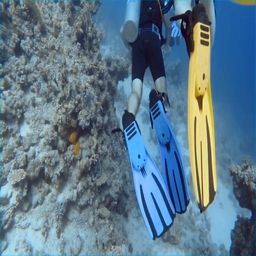}}

			\centerline{(g)}
	\end{subfigure}\hfill
	\begin{subfigure}[b]{0.09\linewidth}
			\centerline{\includegraphics[width=1.2cm,height=1.2cm]{uie/wm/r90/1}}
			\centerline{\includegraphics[width=1.2cm,height=1.2cm]{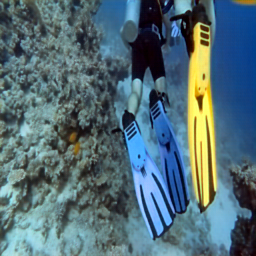}} 
			
			\centerline{(h)}
	\end{subfigure}\hfill
    \begin{subfigure}[b]{0.09\linewidth}
			\centerline{\includegraphics[width=1.2cm,height=1.2cm]{uie/pixmamba/t90/t90_1_25_img_.png}}
			\centerline{\includegraphics[width=1.2cm,height=1.2cm]{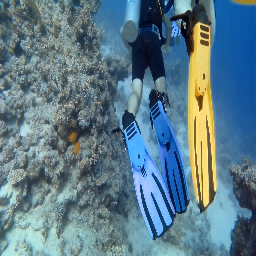}} 
			
			\centerline{(i)}
	\end{subfigure}\hfill
	\begin{subfigure}[b]{0.09\linewidth}
			\centerline{\includegraphics[width=1.2cm,height=1.2cm]{uie/gt/r90/1}}
			\centerline{\includegraphics[width=1.2cm,height=1.2cm]{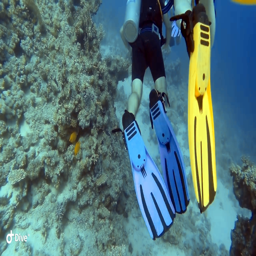}} 
			
			\centerline{(j)}
	\end{subfigure}\hfill
 
          
			
			
			

  \caption{The qualitative comparisons. T90\cite{li2020uieb} samples are presented in each row from top to bottom. (a) raw; (b) Ucolor\cite{li2021ucolor}; (c) PUGAN\cite{cong2023PUGAN}; (d) MFEF\cite{zhou2023mfef}; (e) Semi-UIR\cite{huang2023semiuir}; (f) Convformer\cite{gu2022convformer}; (g) X-CAUNET\cite{pramanick2024xcaunet}; (h) WaterMamba\cite{guan2024watermamba}; (i) PixMamba; (j) reference.}

  \label{fig:qual_t90}
\end{figure}

\subsection{Qualitative Comparison}

\begin{figure}[!tb]    
	\centering
	\begin{subfigure}[b]{0.07\linewidth}
        \centering
        \centerline{\includegraphics[width=1.2cm,height=1.2cm]{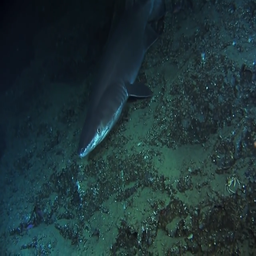}}
        \centerline{\includegraphics[width=1.2cm,height=1.2cm]{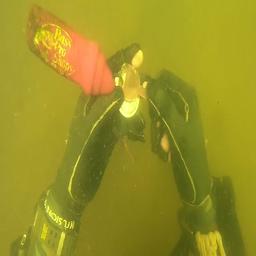}}
        \centerline{\includegraphics[width=1.2cm,height=1.2cm]{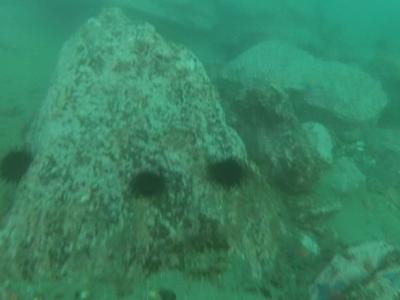}}
        \centerline{(a)}
	\end{subfigure}\hfill
	\begin{subfigure}[b]{0.07\linewidth}
        \centering
        \centerline{\includegraphics[width=1.2cm,height=1.2cm]{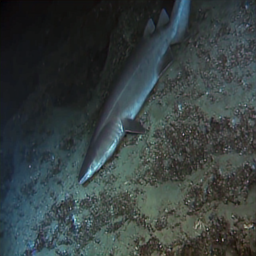}}
        \centerline{\includegraphics[width=1.2cm,height=1.2cm]{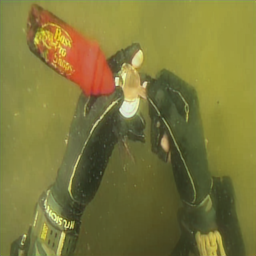}} 
        \centerline{\includegraphics[width=1.2cm,height=1.2cm]{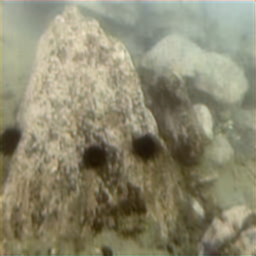}}
        \centerline{(b)}
	\end{subfigure}\hfill
	\begin{subfigure}[b]{0.07\linewidth}
        \centerline{\includegraphics[width=1.2cm,height=1.2cm]{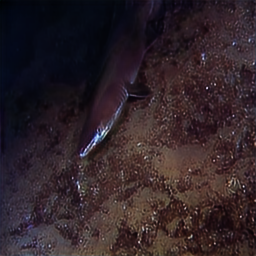}}
        \centerline{\includegraphics[width=1.2cm,height=1.2cm]{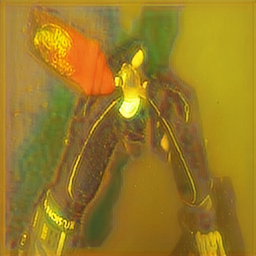}} 
        \centerline{\includegraphics[width=1.2cm,height=1.2cm]{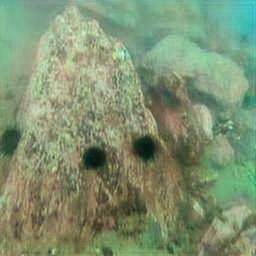}}
     
        \centerline{(e)}
	\end{subfigure}\hfill
	\begin{subfigure}[b]{0.07\linewidth}

        \centerline{\includegraphics[width=1.2cm,height=1.2cm]{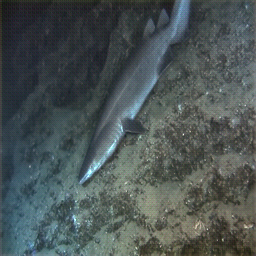}}
        \centerline{\includegraphics[width=1.2cm,height=1.2cm]{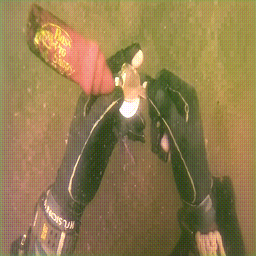}} 
        \centerline{\includegraphics[width=1.2cm,height=1.2cm]{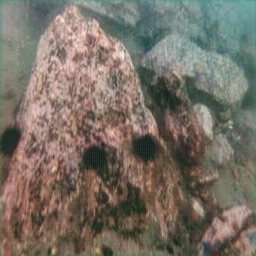}}

        \centerline{(d)}
	\end{subfigure}\hfill
	\begin{subfigure}[b]{0.07\linewidth}

        \centerline{\includegraphics[width=1.2cm,height=1.2cm]{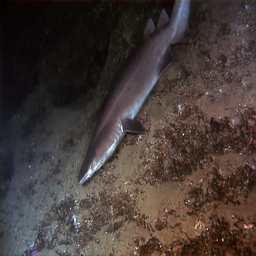}}
        \centerline{\includegraphics[width=1.2cm,height=1.2cm]{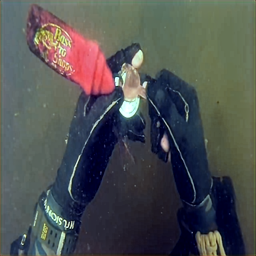}} 
        \centerline{\includegraphics[width=1.2cm,height=1.2cm]{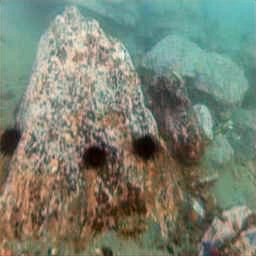}}

        \centerline{(e)}
	\end{subfigure}\hfill
	\begin{subfigure}[b]{0.07\linewidth}

        \centerline{\includegraphics[width=1.2cm,height=1.2cm]{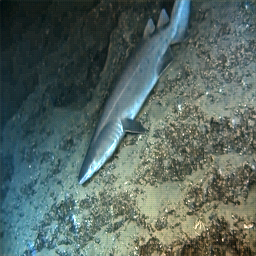}}
        \centerline{\includegraphics[width=1.2cm,height=1.2cm]{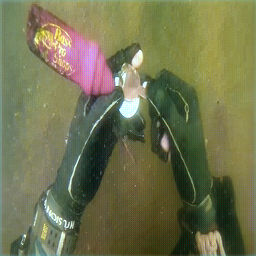}} 
        \centerline{\includegraphics[width=1.2cm,height=1.2cm]{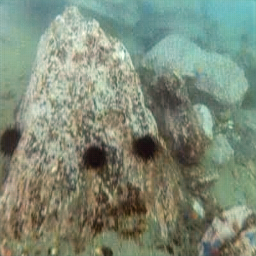}}

        \centerline{(f)}
	\end{subfigure}\hfill
	\begin{subfigure}[b]{0.07\linewidth}
 
        \centerline{\includegraphics[width=1.2cm,height=1.2cm]{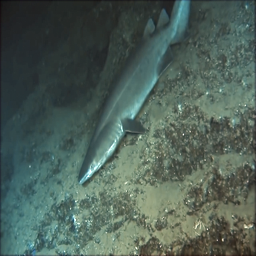}}
        \centerline{\includegraphics[width=1.2cm,height=1.2cm]{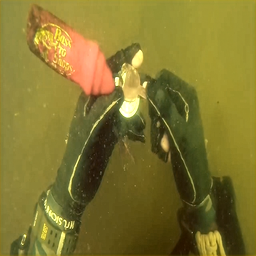}} 
        \centerline{\includegraphics[width=1.2cm,height=1.2cm]{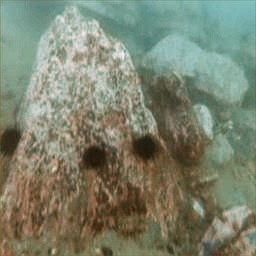}}
 
        \centerline{(g)}
	\end{subfigure}\hfill
	\begin{subfigure}[b]{0.07\linewidth}

        \centerline{\includegraphics[width=1.2cm,height=1.2cm]{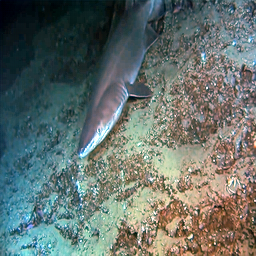}}
        \centerline{\includegraphics[width=1.2cm,height=1.2cm]{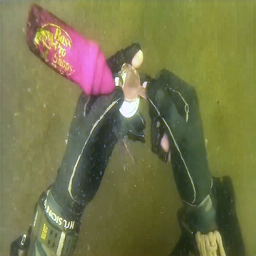}} 
        \centerline{\includegraphics[width=1.2cm,height=1.2cm]{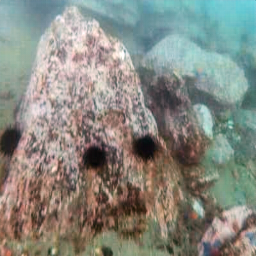}}

        \centerline{(h)}
	\end{subfigure}\hfill
    \begin{subfigure}[b]{0.07\linewidth}

        \centerline{\includegraphics[width=1.2cm,height=1.2cm]{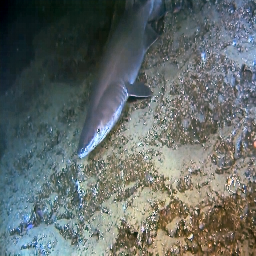}}
        \centerline{\includegraphics[width=1.2cm,height=1.2cm]{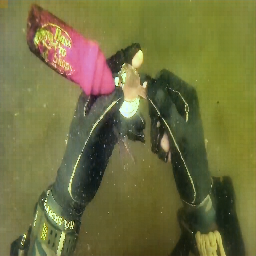}} 
        \centerline{\includegraphics[width=1.2cm,height=1.2cm]{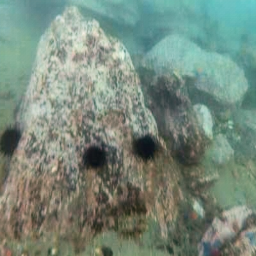}}

        \centerline{(i)}
	\end{subfigure}\hfill

  \caption{The qualitative comparisons. First and second row are C60\cite{li2020uieb} samples. Third row is  UCCS\cite{liu2020uccs} samples. (a) raw; (b) Ucolor\cite{li2021ucolor}; (c) PUGAN\cite{cong2023PUGAN}; (d) MFEF\cite{zhou2023mfef}; (e) Semi-UIR\cite{huang2023semiuir}; (f) Convformer\cite{gu2022convformer}; (g) X-CAUNET\cite{pramanick2024xcaunet}; (h) WaterMamba\cite{guan2024watermamba}; (i) PixMamba.}


          

      



  \label{fig:qual_c60_uccs}
\end{figure}

The visual qualitative comparison of our proposed PixMamba and other state-of-the-art models is depicted in \cref{fig:qual_t90} and \cref{fig:qual_c60_uccs}. We reported most representative samples from the datasets. Additional, we illustrated the detail comparison in \cref{fig:detail_comp}. By zooming in on fine-grained details, PixMamba enhances the entire degraded underwater image while preserving the quality of image. This advancement enables UIE to be further applied in high-resolution scenarios.

\subsection{Quantitative Comparisons}

As shown in \cref{table_c60_uccs}, we compare our PixMamba with several state-of-the-art models included UColor\cite{li2021ucolor}, UIEC\^{}2-Net\cite{wang2021uiec}, Shallow-UWNet\cite{naik2021shallowuwnet}, PUIE-Net\cite{fu2022puienet}, URSCT\cite{ren2022urstc}, Restormer\cite{zamir2021restormer}, PUGAN\cite{cong2023PUGAN}, MFEF\cite{zhou2023mfef}, Semi-UIR\cite{huang2023semiuir}, Convformer\cite{gu2022convformer}, X-CAUNET\cite{pramanick2024xcaunet}, Five A$^{+}$\cite{jiang2023five}, NU$^2$Net\cite{guo2023uranker}, and WaterMamba\cite{guan2024watermamba}. The proposed PixMamba outperforms other state-of-the-art models across various datasets. Compared to Semi-UIR\cite{huang2023semiuir}, our method has improved UIQM and UCIQE by 0.201 and 0.012 on C60 dataset, and improved UCIQE by 0.007 on UCCS datasets. On T90 dataset, compared to NU$^{2}$Net\cite{guo2023uranker}, the PSNR, UIQM, and UCIQE were improved by 0.526, 0.112, and 0.03, respectively. 

\subsection{Ablation Studies}
\begin{table}[h]
	\centering
 \caption{Ablation study. Trained on \textbf{U800} dataset and validated on \textbf{T90} dataset.}
	\fontsize{7}{11}\selectfont
	\setlength{\tabcolsep}{0.8mm}{
	\begin{tabular}{cccccccccccccc}
		\hline
    		 \multirow{2}{*}{ \textbf{Method}} & \multirow{2}{*}{\textbf{EMNet}} & \multirow{2}{*}{\textbf{MUB}} & \multirow{2}{*}{\textbf{PixNet}} & \multirow{2}{*}{\textbf{BPE}} & \multicolumn{2}{c}{\textbf{T90}} & \multirow{2}{*}{\textbf{Params} $\downarrow$} & \multirow{2}{*}{\textbf{FLOPs} $\downarrow$}
        \\ &&&&&PSNR $\uparrow$&SSIM $\uparrow$ \\

        \hline
        U-Net (ResBlock) \cite{ronneberger2015unet} & & & & & 18.102 & 0.822 & 3.35M & 17.42G \\
		\hline
        PixMamba & \checkmark &            &            &            & 22.857 & 0.913 & 7.05M & 7.15G \\
        PixMamba & \checkmark & \checkmark &            &            & 22.969 & 0.919 & 8.66M & 5.99G \\
        PixMamba & \checkmark & \checkmark & \checkmark &            & 23.295 & 0.920 & 8.68M & 7.60G \\
        PixMamba & \checkmark & \checkmark & \checkmark & \checkmark & 23.587 & 0.921 & 8.68M & 7.60G \\
        \hline
		\end{tabular}
	}
    \label{tab:ablation}
\end{table}

To evaluate the contributions of each component in our proposed PixMamba model, we conduct an ablation study, summarized in \cref{tab:ablation}. U-Net (ResBlock) \cite{ronneberger2015unet}, reaches 18.102 PSNR and 0.822 SSIM with 3.35M parameters and 17.42G FLOPs. Introducing the Efficient Mamba Net (EMNet) module and replaced the Mamba Upsampling Block (MUB) with vanilla patch expand upsampling\cite{hu2022swinunet} to form the initial PixMamba architecture boosts the PSNR to 22.857 and the SSIM to 0.913, albeit with a higher parameter count of 7.05M and reduced FLOPs of 7.15G. Adding the Mamba Upsampling Block (MUB) to PixMamba further improves the PSNR to 22.969 and SSIM to 0.919, though it increases the parameter count to 8.66M while reducing the FLOPs to 5.99G. Adding the PixMamba Net (PixNet) without integrates our proposed Block-wise Positional Embedding (BPE) improves the PSNR to 23.29 and SSIM to 0.920, with a slight increase in parameters to 8.68M and FLOPs to 7.60G. Finally, incorporating the PixMamba Net (PixNet) into the architecture enhances the performance further, achieving a PSNR of 23.587 and an SSIM of 0.921. This comprehensive analysis demonstrates that each module in the PixMamba architecture incrementally contributes to the overall performance.

\section{Conclusion}

In this paper, we introduced a novel architecture for underwater image enhancement (UIE) task: PixMamba, which leverages State Space Models (SSM) for linear complexity and effective feature modeling. PixMamba employs a dual-level processing approach, which contains Efficient Mamba Net (EMNet) for patch-level modeling and PixMamba Net (PixNet) for pixel-level modeling to improve overall image quality and model efficiency. Specially, PixNet contains Block-wise Positional Embedding (BPE) while modeling at pixel-level patch, it allows PixNet to have both spatial information and global fine-grained features seamlessly. EMNet utilizes an SSM-based U-Net architecture at the patch level. It incorporates two key components: Efficient Mamba Block (EMB) for lower memory computational cost and Mamba Upsampling Block (MUB) for more detail-preserving restoration. Comprehensive experiments demonstrate that PixMamba performs advantageously against existing methods, substantiating its efficiency and effectiveness.

\noindent \textbf{Acknowledgements.}
No acknowledgements.

\bibliographystyle{splncs04}
\bibliography{ref}

\appendix

\end{document}